\documentclass[10pt,twocolumn,letterpaper]{article}

\usepackage{cvpr}              % To produce the CAMERA-READY version
\usepackage{graphicx}
\usepackage{amsmath}
\usepackage{amssymb}
\usepackage{booktabs}

\usepackage{microtype}
\usepackage{booktabs} % for professional tables
\usepackage{multirow}
\usepackage{multicol}
\usepackage{color, colortbl}
\usepackage{adjustbox}
\usepackage{xspace}
\usepackage[pagebackref,breaklinks,colorlinks]{hyperref}

\usepackage[capitalize]{cleveref}
\crefname{section}{Sec.}{Secs.}
\Crefname{section}{Section}{Sections}
\Crefname{table}{Table}{Tables}
\crefname{table}{Tab.}{Tabs.}

\def\cvprPaperID{5619} % *** Enter the CVPR Paper ID here
\def\confName{CVPR}
\def\confYear{2022}

\begin{document}

%%%%%%%%% TITLE - PLEASE UPDATE
\title{Few-Shot Object Detection with Fully Cross-Transformer}

\author{Guangxing Han, Jiawei Ma, Shiyuan Huang, Long Chen, Shih-Fu Chang\\
Columbia University\\
{
\tt\small \{gh2561,jiawei.m,sh3813,cl3695,sc250\}@columbia.edu}
}

\maketitle

%%%%%%%%% ABSTRACT
\begin{abstract}
Few-shot object detection (FSOD), with the aim to detect novel objects using very few training examples, has recently attracted great research interest in the community. Metric-learning based methods have been demonstrated to be effective for this task using a two-branch based siamese network, and calculate the similarity between image regions and few-shot examples for detection. However, in previous works, the interaction between the two branches is only restricted in the detection head, while leaving the remaining hundreds of layers for separate feature extraction. Inspired by the recent work on vision transformers and vision-language transformers, we propose a novel \textbf{F}ully \textbf{C}ross-\textbf{T}ransformer based model (FCT) for FSOD by incorporating cross-transformer into both the feature backbone and detection head. The asymmetric-batched cross-attention is proposed to aggregate the key information from the two branches with different batch sizes. Our model can improve the few-shot similarity learning between the two branches by introducing the multi-level interactions. Comprehensive experiments on both PASCAL VOC and MSCOCO FSOD benchmarks demonstrate the effectiveness of our model. 
\end{abstract}

%%%%%%%%% BODY TEXT
\section{Introduction}

Few-shot object detection (FSOD) aims to detect objects from the query image using a few training examples. This is motivated by human visual system which can quickly learn novel concepts from very few instructions. The key point is how to quickly learn object detection models with strong generalization ability using a small number of training data, such that the learned model can detect objects in unseen images.
This is very challenging, especially for the current state-of-the-art deep-learning based methods \cite{ren2015faster,redmon2016you,liu2016ssd,carion2020end}, which usually need thousands of training examples and are prone to overfitting under this data-scarce scenario.

\begin{figure}[t]
\begin{center}
\includegraphics[scale=0.35]{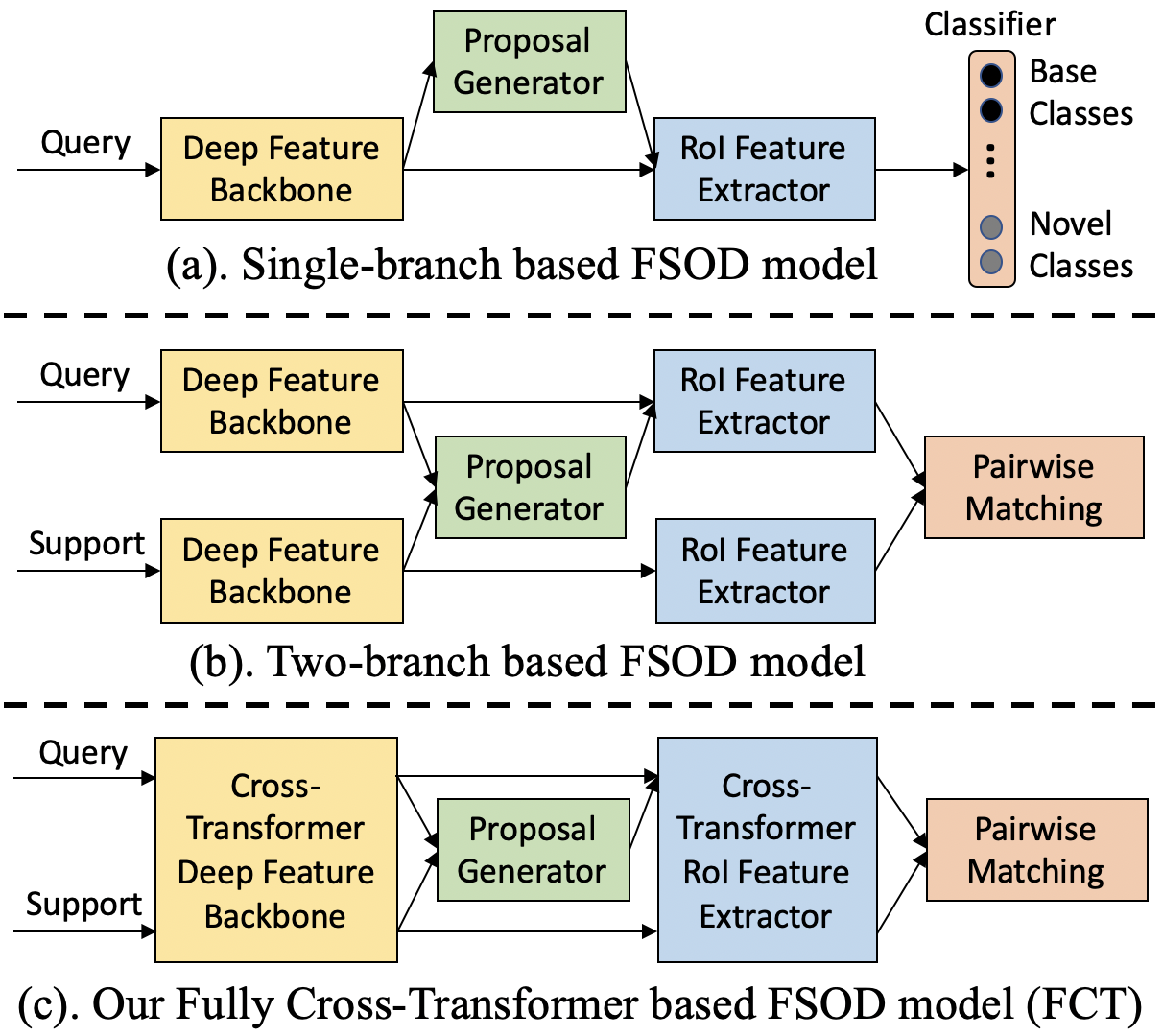}
\end{center}
\vspace{-2mm}
\caption{Comparison of the single-branch, two-branch based FSOD models and our proposed model.}
\vspace{-2mm}
\label{figure_1}
\end{figure}

Current methods for this task mainly follow a two-stage learning paradigm \cite{wang2020few} to transfer the knowledge learned from the data-abundant base classes to assist in object detection for few-shot novel classes.
The detailed model architectures vary in different works, which can be roughly divided into two categories, single-branch based methods \cite{wang2020few,wu2020multi,Sun_2021_CVPR,Zhang_2021_CVPR,Zhu_2021_CVPR} and two-branch based methods \cite{kang2019few,fan2020few,yan2019meta,han2021meta,Han_2021_ICCV,hsieh2019one}.
(1) Single-branch based methods employ a typical object detection model, \eg, Faster R-CNN \cite{ren2015faster}, and build a multi-class classifier for detection. It is prone to overfitting to the small training data, especially when we only have 1-shot training data per novel class.
(2) Two-branch based methods apply the metric-learning idea \cite{vinyals2016matching,snell2017prototypical,sung2018learning} to FSOD and build a siamese network to process the query image and the few-shot support image in parallel. After extracting deep visual features from the two branches, previous works propose various methods (\eg, feature fusion \cite{fan2020few,yan2019meta,xiao2020few}, feature alignment \cite{han2021meta}, GCN \cite{Han_2021_ICCV}, and non-local attention/transformer \cite{wang2018non,hsieh2019one,chen2021dual,NEURIPS2020_fa28c6cd,chen2021adaptive}) to calculate the similarity of the two branches. 
The two-branch based methods do not learn the multi-class classifier over novel classes, and usually have stronger generalization ability by learning to compare the query regions with few-shot classes.

Previous two-branch based methods have explored various interactions (\eg, alignment) between the query and support branch to improve the similarity learning. But the interactions are restricted in the detection head with high-level features, and leave the remaining hundreds of layers for separate feature extraction. In fact, the query and support images may have large visual differences and domain gap in terms of object pose, scale, illumination, occlusion, background and etc. Simply aligning the two branches at the high-level feature space might not be optimal. If we could align the extracted features in all network layers, the network could have more capacity focusing on the common features in each layer, and improve the final similarity learning.

In this work, we propose a novel Fully Cross-Transformer based model (FCT) for FSOD, which is a pure cross-transformer based detection model without deep convolutional networks. The ability to model long-range dependencies in transformer \cite{vaswani2017attention} can not only capture the abundant context in one branch, and also related context in the other branch, thus encouraging mutual alignment between the two branches.
As shown in Figure \ref{figure_1}, Our model is based on the two-stage detection model Faster R-CNN. Instead of extracting deep visual features separately for the query and support inputs, we use the multi-layer deep cross-transformer to jointly extract features for the two branches. 
Inside the cross-transformer layer, we propose the asymmetric-batched cross-attention to aggregate the key information from the two branches with different batch sizes, and update the features of either branch using self-attention with the aggregated key information. Thus, we can align the features from the two branches in each of the cross-transformer layer.
Then after the joint feature extraction and proposal generation for the query image, we propose a cross-transformer based RoI feature extractor in the detection head to jointly extract RoI features for the query proposals and support images. Incorporating our cross-transformer in both feature backbone and ROI feature extractor could largely promote the multi-level interactions (alignment) between the query and support inputs, thus further improving the final FSOD performance.

We'd like to emphasize the difference between a closely related work ViLT \cite{vilt} and ours, both using transformers for joint feature extraction of two branches. \textbf{First}, ViLT has language and the original image as input, and the highly abstracted language tokens are interacting with the visual tokens at each layer. However, visual tokens represent low-level concepts at the beginning, and evolve into high-level concepts in deep layers.
Different from ViLT, we take input of two visual images, and explore multi-level interactions between the two visual branches, gradually from low-level to high-level features. \textbf{Second}, we focus on FSOD, a dense prediction task, instead of the classification and retrieval task in ViLT, and incorporate cross-transformer into both the feature backbone and detection head. \textbf{Third}, ViLT extracts visual tokens following ViT \cite{dosovitskiy2020vit}, and uses the same number of tokens throughout the model. We employ the pyramid structure \cite{Wang_2021_ICCV} to extract multi-scale visual tokens, and propose the asymmetric-batched cross-attention across the branches with different batch sizes to reduce computational complexity.

Our contributions can be summarized as:
   (1) To the best of our knowledge, we are the first to explore and propose the vision transformer based few-shot object detection model.
   (2) A novel fully cross-transformer is proposed for both the feature backbone and detection head, to encourage multi-level interactions between the query and support. We also propose the asymmetric-batched cross-attention across the branches.
   (3) We comprehensively evaluate the proposed model on the two widely used FSOD benchmarks and achieve state-of-the-art performance.

\section{Related Works}

\textbf{Object Detection.} Object detection is one of the most fundamental tasks in computer vision. Recently, deep convolutional neural networks (DCNNs \cite{ALEXNET,he2016deep}) have demonstrated their power to automatically learn feature from a large scale of training data, and are the dominant approach for object detection. Current methods using DCNNs can mainly be grouped into two categories: proposal-based methods and proposal-free methods. Proposal-based methods \cite{Fast_R-CNN,ren2015faster,he2017mask,R_RPN,han2018semi} divide object detection into two sequential stages by firstly generating a set of region proposals and then performing classification and bounding box regression for each proposal. Proposal-free methods \cite{redmon2016you,liu2016ssd,tian2019fcos,SSD_TDR} directly predict the bounding boxes and the corresponding class labels on top of CNN features. Recently, the transformer based object detection models \cite{carion2020end,zhu2020deformable} show promising results, but still suffer from slow convergence problem. Therefore, we choose to use one of the most representative proposal-based methods, Faster R-CNN \cite{ren2015faster}, for FSOD considering both detection accuracy and training efficiency.

\textbf{Few-Shot Learning.} Few-shot learning (FSL) aims to recognize novel classes using only few examples.
The key idea of FSL is to transfer knowledge from many-shot base classes to few-shot novel classes. Existing few-shot learning methods can be roughly divided into the following three categories: (1) Optimization based methods. For example, model-Agnostic Meta-Learning (MAML \cite{finn2017model}) learns a good initialization so that the learner could rapidly adapt to novel tasks within a few optimization steps. (2) Parameter generation based methods \cite{gidaris2018dynamic,Huang_2022_CVPR}. For example, \textit{Gidaris et al.} \cite{gidaris2018dynamic} proposes an attention-based weight generator to generate the classifier weights for novel classes. (3) Metric-learning based methods \cite{vinyals2016matching,snell2017prototypical,sung2018learning,Ma_2021_ICCV,ypsilantis2021met}. These methods learn a generalizable similarity metric-space from base classes. For example, Prototypical Networks \cite{snell2017prototypical} calculate prototype of novel classes by averaging the features of the few samples, and then perform classification by a nearest neighbor search.

\textbf{Few-Shot Object Detection.}
Few-shot object detection needs to not only recognize novel objects using a few training examples, but also localize objects in the image.
Existing works can be mainly grouped into the following two categories according to the model architecture: (1) Single-branch based methods \cite{wang2020few,wu2020multi,Sun_2021_CVPR,Zhang_2021_CVPR,Zhu_2021_CVPR}. These methods attempt to learn object detection using the long-tailed training data from both data-abundant base classes and data-scarce novel classes. The final classification layer in the detection head is determined by the number of classes to detect. To deal with the unbalanced training set, re-sampling \cite{wang2020few} and re-weighting \cite{lin2017focal} are the two main strategies. \textit{Wang et al.} \cite{wang2020few} shows that a simple two-stage fine-tuning approach outperforms other complex meta-learning methods. 
Following works introduce multi-scale positive sample refinement \cite{wu2020multi}, image hallucination \cite{Zhang_2021_CVPR}, contrastive learning \cite{Sun_2021_CVPR} and linguistic semantic knowledge \cite{Zhu_2021_CVPR} to assist in FSOD.
(2) Two-branch based methods \cite{kang2019few,fan2020few,yan2019meta,han2021meta,Han_2021_ICCV,Han_2022_arXiv,hsieh2019one}.
These methods are based on a siamese network to process the query and support in parallel, and calculate the similarity between image regions (usually proposals) and few-shot examples for detection. \textit{Kang et al.} \cite{kang2019few} first propose a feature reweighting module to aggregate the query and support features. Multiple feature fusion networks \cite{fan2020few,yan2019meta,han2021meta,xiao2020few} are then proposed for stronger feature aggregation. \textit{Han et al.} \cite{han2021meta} propose to perform feature alignment between the two inputs and focus on foreground regions using attention. GCNs are employed in \cite{Han_2021_ICCV} to facilitate mutual adaptation between the two branches. Other works \cite{hsieh2019one,chen2021dual,NEURIPS2020_fa28c6cd,chen2021adaptive} use more advanced non-local attention/transformer \cite{wang2018non,vaswani2017attention} to improve the similarity learning of the two inputs.
All these previous works show that the two-branch paradigm is a promising solution for FSOD. Our work also belongs to this category, and proposes a pure cross-transformer model to exploit the interaction between the two branches to the largest extent.

\textbf{Transformer and Its Application in Computer Vision.} Transformer was first introduced by \textit{Vaswani et al.} \cite{vaswani2017attention} as a new attention-based building block for machine translation and has become a prevalent architecture in NLP \cite{devlin2018bert}. The success of transformer can be attributed to its strong ability to model long-range dependencies using self-attention. Since then, transformer has been extended to various vision-related tasks, \eg, vision-and-language pre-training \cite{Su2020VL-BERT:,vilt,tan2019lxmert}, image classification \cite{dosovitskiy2020vit,Wang_2021_ICCV,liu2021Swin}, object detection \cite{carion2020end,zhu2020deformable}, and etc. The pioneering work of Vision Transformer (ViT \cite{dosovitskiy2020vit}) splits an image into non-overlapping patches (similar to tokens in NLP) and provides the sequence of linear embeddings of these patches as an input to a transformer, and show promising results for image classification compared with CNNs \cite{he2016deep}. Following works \eg, PVT \cite{Wang_2021_ICCV,wang2021pvtv2}, Swin \cite{liu2021Swin}, and Twins \cite{chu2021Twins}, introduce pyramid structure to generate multi-scale feature maps for dense prediction tasks. Spatial-reduction attention \cite{Wang_2021_ICCV,wang2021pvtv2} and Shifted Window based Self-Attention \cite{liu2021Swin} are proposed to reduce the computational complexity in the transformer. 
\textit{Kim et al.} \cite{vilt} propose a unified vision-language transformer model without convolution (ViLT \cite{vilt}), to focus more on the modality interactions instead of using deep modal-specific embeddings.
Our work is inspired by these previous works, and propose a novel fully cross-transformer based FSOD model.

\section{Our Approach}

\subsection{Task Definition}
In few-shot object detection (FSOD), we have two sets of classes $C = C_{base} \cup C_{novel}$ and $C_{base} \cap C_{novel} = \emptyset$, where base classes $C_{base}$ have plenty of training data per class, and novel classes $C_{novel}$ (a.k.a. support classes) only have very few training examples for each class (a.k.a. support images). For $K$-shot (\eg, $K=1,5,10$) object detection, we have exactly $K$ bounding box annotations for each novel class $c \in C_{novel}$ as the training data. The goal of FSOD is to leverage the data-abundant base classes to assist in detection for few-shot novel classes. 

\begin{figure*}[t]
\begin{center}
\includegraphics[scale=0.4]{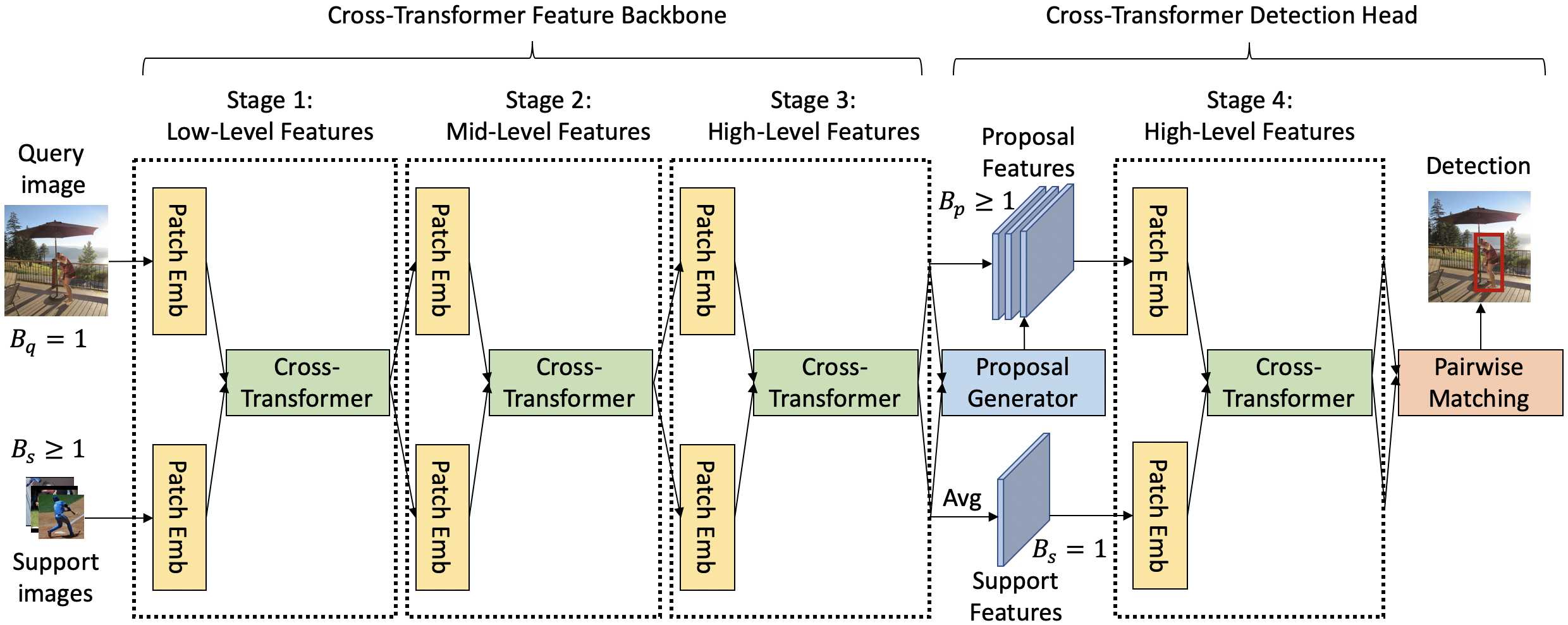}
\end{center}
\vspace{-2mm}
\caption{The overall architecture of our proposed Fully Cross-Transformer based few-shot object detection model (FCT).}
\vspace{-2mm}
\label{figure_2}
\end{figure*}

\subsection{Overview of Our Proposed Model (FCT)}

We propose a novel Fully Cross-Transformer (FCT) based few-shot object detection model in this work. Our work belongs to the two-branch based few-shot object detection method. The motivation is that although the traditional two-branch based methods \cite{kang2019few,fan2020few,yan2019meta,han2021meta,Han_2021_ICCV,hsieh2019one} show promising results, the interaction of the query and support branch is only restricted in the detection head, while leaving hundreds of layers for separate feature extraction in each branch before the cross-branch interaction. Our idea is to remove the separate deep feature encoders and fully exploit the cross-branch interaction to the largest extend.

An overview of our model is illustrated in Figure \ref{figure_2}. Our model is based on the Faster R-CNN object detection framework. In Faster R-CNN, we have a feature backbone to extract deep visual features of the input. Then proposals are generated using the extracted features and a detection head is followed to extract the RoI features for each proposal and perform classification and bounding box (bbox) refinement. Inspired by the recent vision transformers and vision-language transformers, we propose a pure cross-transformer based few-shot object detection model without deep convolutional networks. Specifically, the cross-transformer is incorporated into both the feature backbone and detection head. We show in Section \ref{backbone} how we jointly extract features for both the query and support images using our cross-transformer feature backbone, and similarly in Section \ref{detection_head} we show the details of our cross-transformer detection head. The model training framework is introduced in Section \ref{training}.

\subsection{The Cross-Transformer Feature Backbone}
\label{backbone}

We have three stages of cross-transformer modules in the feature backbone for joint feature extraction of the query and support inputs. In the first stage, we have a single query image $I_q \in \mathbb{R}^{1*H_{I_q}*W_{I_q}*3}$ and a batch of support images $I_s \in \mathbb{R}^{B_s*H_{I_s}*W_{I_s}*3}$ of the same class as inputs, where $B_s \geq 1$. We first split the original RGB images into non-overlapping $4\times4\times3$ patches. Then the flattened patches go through a linear patching embedding layer and are projected to $C_1$ dimensions. The embedded patch sequences $X_q \in \mathbb{R}^{N_1^q*C_1}$ ($N_1^q=\frac{H_{I_q}}{4}*\frac{W_{I_q}}{4}$) and $X_s \in \mathbb{R}^{N_1^s*C_1}$ ($N_1^s=\frac{H_{I_s}}{4}*\frac{W_{I_s}}{4}$) of the two branches are fed into several cross-transformer layers. The second and third stage share a similar architecture as the first stage, and generate feature maps with gradually decreasing sequence lengths and increasing channel dimensions.

Following the vallina transformer \cite{vaswani2017attention}, our cross-transformer layer consists of the proposed multi-head asymmetric-batched cross-attention and two feed-forward layers, with LayerNorm (LN), GELU non-linearity and residual connections in between. 

Specifically, the position embedding $\mathbf{E}_{q}^{pos} \in \mathbb{R}^{N_1^q*C_1}$, $\mathbf{E}_{s}^{pos} \in \mathbb{R}^{N_1^s*C_1}$ and branch embedding $\mathbf{E}^{bra} \in \mathbb{R}^{2*C_1}$ are first added to the input patch sequences $X_q$ and $X_s$ to retain the position and branch information,
\begin{equation}
X_q^{'} = X_q + \mathbf{E}_{q}^{pos} + \mathbf{E}^{bra}[0],\ X_s^{'} = X_s + \mathbf{E}_{s}^{pos} + \mathbf{E}^{bra}[1] 
\end{equation}

In multi-head cross-attention, we map the input patch sequence $X_q^{'}$ to $Q_q^{i}, K_q^{i}, V_q^{i}$ and $X_s^{'}$ to $Q_s^{i}, K_s^{i}, V_s^{i}$ in the head $i$ ($i=1...h$, and $h$ is the the number of head), following the Q-K-V attention in transformer \cite{vaswani2017attention}. In order to reduce the computational complexity of the attention, especially in the early layers, inspired by PVT \cite{Wang_2021_ICCV}, we use the spatial-reduction operation to sub-sample the feature maps for K and V. Another benefit is that we can summarize the key information using the sub-sampled K and V,
\begin{gather}
Q_q^i = X_q^{'}W_Q^i, \quad Q_s^i = X_s^{'}W_Q^i \\
K_q^i = \mathbf{SR}(X_q^{'})W_K^i, \quad K_s^i = \mathbf{SR}(X_s^{'})W_K^i \\
V_q^i = \mathbf{SR}(X_q^{'})W_V^i, \quad V_s^i = \mathbf{SR}(X_s^{'})W_V^i
\end{gather}
\noindent where $W_Q^i \in \mathbb{R}^{C_1*d_{h}}, W_K^i \in \mathbb{R}^{C_1*d_{h}}, W_V^i \in \mathbb{R}^{C_1*d_{h}}$ are the learnable weights of the linear projection, which are shared between the two branches. The dimension of the projected features is $d_{h} = C_1/h$, same in each head. $\mathbf{SR}(\cdot)$ is the spatial-reduction operation, and can be implemented by a strided convolution layer or a spatial pooling layer.

\paragraph{The Asymmetric-Batched Cross-Attention.} The batch size of the query branch and support branch is different. We perform detection for each query image separately, because different query images are irrelevant and the detection is independent from each other. For the support branch, the novel classes are also processed one-by-one, but the number of support images for one class could be arbitrary. The naive implementation of only forwarding a query image and a support image each time and repeating the process for each support image can be extremely slow. Therefore, we propose the asymmetric-batched cross-attention to calculate the attention between the query image and all support images of the same class at one time. 

\begin{figure}[t]
\begin{center}
\includegraphics[scale=0.36]{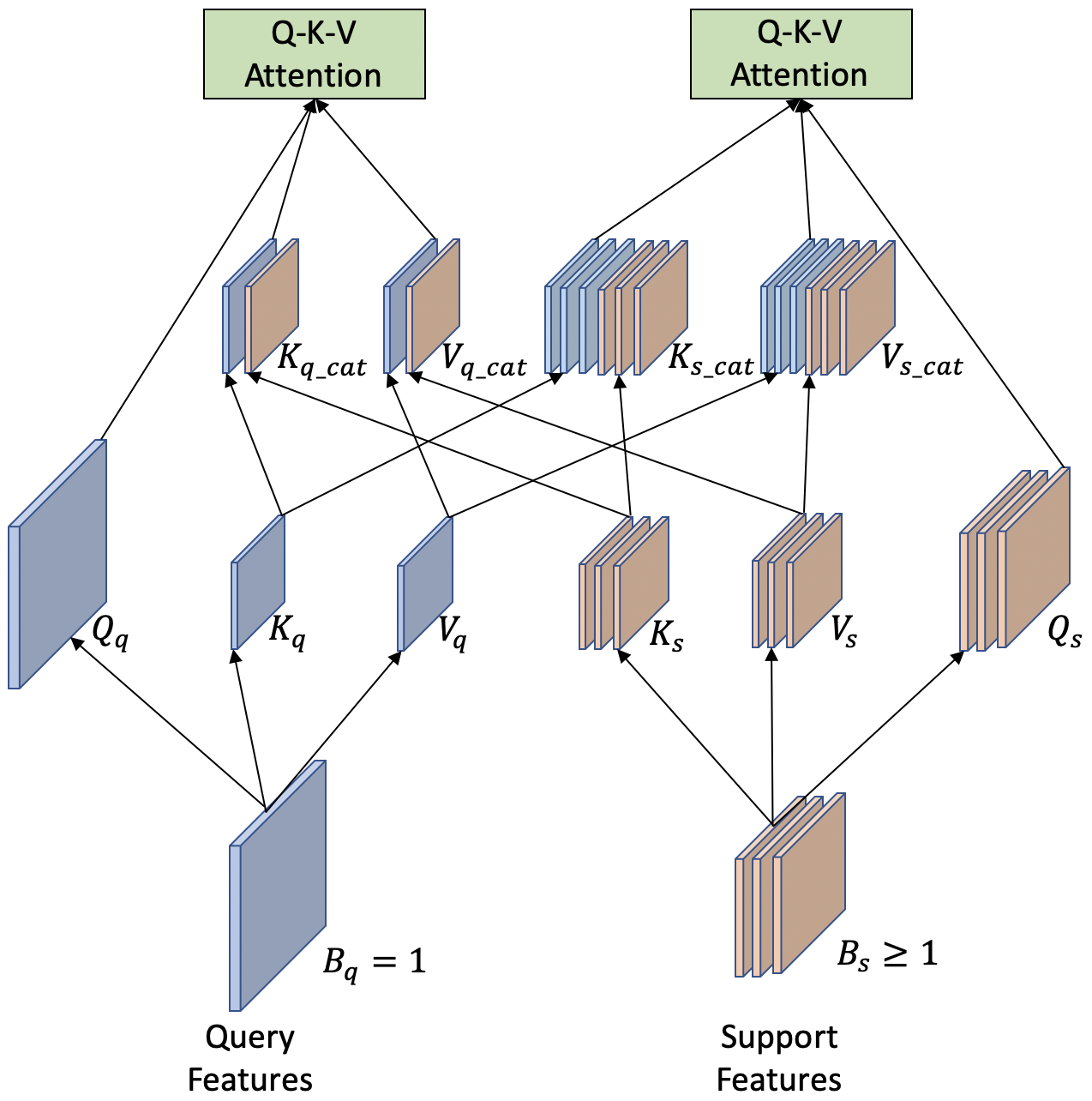}
\end{center}
\vspace{-4mm}
\caption{The proposed Asymmetric-Batched Cross-Attention in our cross-transformer feature backbone.}
\vspace{-2mm}
\label{figure_3}
\end{figure}

As shown in Figure \ref{figure_3}, the cross-attention layer aggregates the key information (K-V pairs) from the two branches for attention. To aggregate the K-V pairs from the support branch to the query branch, we first conduct average pooling over the multiple support images to match the batch size of the query branch, and then concatenate the K-V pairs of the two branches. Similarly, to aggregate the K-V pairs from the query branch to the support branch, we first repeat the query image $B_s$ times along the batch dimension, and then concatenate the K-V pairs of the two branches,
\begin{align}
K_{q\_{cat}}^{i} &= [K_q^i, \frac{1}{B_s}\textstyle{\sum}_{B_s}K_s^i], \\ 
V_{q\_{cat}}^{i} &= [V_q^i, \frac{1}{B_s}\textstyle{\sum}_{B_s}V_s^i] \\
K_{s\_{cat}}^{i} &= [\mathbf{REP}(K_q^i, B_s), K_s^i], \\ 
V_{s\_{cat}}^{i} &= [\mathbf{REP}(V_q^i, B_s), V_s^i],
\end{align}
where $[\cdot,\cdot]$ denotes the concatenation along the token dimension by default, and $\mathbf{REP}(A, b)$ is to repeat the tensor $A$ by $b$ times along the batch dimension by default.

Thus, the multi-head asymmetric-batched cross-attention can be summarized as,
\begin{align}
X_q^{''} &= \mathrm{Concat}(\mathrm{head_q^1}, ..., \mathrm{head_q^h})W_O \\
\mathrm{where}\ \mathrm{head_s^i} &= \mathrm{Attention}(Q_q^i, K_{q\_{cat}}^i, V_{q\_{cat}}^i) \\
X_s^{''} &= \mathrm{Concat}(\mathrm{head_s^1}, ..., \mathrm{head_s^h})W_O \\
\mathrm{where}\ \mathrm{head_s^i} &= \mathrm{Attention}(Q_s^i, K_{s\_{cat}}^i, V_{s\_{cat}}^i)
\end{align}
where $W_O \in \mathbb{R}^{hd_{h}*C_1}$ is the weight of the projection back to the original feature space, shared with the two branches.

\paragraph{} 
Then the feed-forward network is applied to each patch with stronger feature representations, following \cite{vaswani2017attention},
\begin{gather}
X_q^{'''} = \mathrm{MLP}(\mathrm{LN}(X_q^{''}) + X_q^{''} \\
X_s^{'''} = \mathrm{MLP}(\mathrm{LN}(X_s^{''}) + X_s^{''}
\end{gather}

\paragraph{Remarks.} We thoroughly study the multi-level interactions between the two visual branches in our proposed model. The three stages in our cross-transformer feature backbone enable efficient interactions of the two branches with low-level, mid-level and high-level visual features gradually.

\subsection{The Cross-Transformer Detection Head}
\label{detection_head}

In the detection head, we first follow the previous work \cite{fan2020few} to generate class-specific proposals in the query image, and use RoIAlign \cite{he2017mask} to extract the initial RoI features for each proposal $f_p \in \mathbb{R}^{B_p*H^{'}*W^{'}*C_3}$, and similarly for the support branch $f_s \in \mathbb{R}^{B_s*H^{'}*W^{'}*C_3}$. ($B_p=100$ by default, and $H^{'}=W^{'}=14$, the default spatial size after RoIAlign.)

Then the RoI feature extractor, also Stage 4 of our cross-transformer, jointly extracts the RoI features for the proposals and support images before the final detection. In order to reduce the computational complexity, we take the average of all support images $f_s^{'}=\frac{1}{B_s}\sum_{B_s}f_s$, such that $f_s^{'} \in \mathbb{R}^{1*H^{'}*W^{'}*C_3}$. We use the proposed asymmetric-batched cross-attention to calculate the attention of the two branches $f_p$ and $f_s^{'}$, similarly in the feature backbone. The difference is that the batch size of the query proposals is $B_p \geq 1$ and $B_s^{'}=1$ for the support branch, which is the reverse in the backbone.

After the joint RoI feature extraction, we use the pairwise matching network in \cite{fan2020few} for the final detection. Binary cross-entropy loss and bbox regression loss are employed for training, following \cite{fan2020few}.

\paragraph{Remarks.} We follow the vallina Faster R-CNN object detection framework, and do not use FPN \cite{lin2017feature} in our model. We find that using FPN does not improve the performance, especially for the two-branch based FSOD methods \cite{fan2020few,yan2019meta,xiao2020few,Han_2021_ICCV,chen2021dual,hsieh2019one}. The cross-transformer based RoI feature extractor in the detection head can encourage mutual alignment between the query proposals and support images, which is crucial for the final pairwise matching.

\subsection{The Model Training Framework}
\label{training}

We have three steps for model training.

\textbf{Pretraining the single-branch based model over base classes.} In the first step, we pretrain our model without using the cross-transformer. Specifically, we use the vallina Faster R-CNN model with the vision transformer backbone \cite{Wang_2021_ICCV,wang2021pvtv2}, and only train the model using the base-class dataset.

\textbf{Training the two-branch based model over base classes.} Then we train the proposed two-branch based model with fully cross-transformer using the base-class dataset, initialized by the pretrained model in the first step. Our proposed FCT model can reuse most of the parameters of the model learned in the first step. The good initialization point in the first step can ease the training of our FCT model.

\textbf{Fine-tuning the two-branch based model over novel classes.} Finally, we fine-tune our FCT model on a sub-sampled dataset of base and novel classes with $K$-shot samples per class, following the previous works \cite{wang2020few,fan2020few}. Fine-tuning could largely improve the adaptation of our model for novel classes by seeing a few examples during training.

\section{Experimental Results}

\subsection{Datasets}

We evaluate our model on two widely-used few-shot object detection benchmarks as follows. 

\textbf{PASCAL VOC.} Following previous works in \cite{kang2019few,wang2020few}, we have three random partitions of base and novel categories. In each partition, the 20 PASCAL VOC categories are split into 15 base classes and 5 novel classes similarly. 
We sample the few-shot images following \cite{wang2020few,Sun_2021_CVPR}, and report AP50 results under shots 1, 2, 3, 5, and 10. We report both single run results using the exact same few-shot images as \cite{wang2020few,kang2019few} and the average results of multiple runs.

\textbf{MSCOCO.} We use the 20 PASCAL VOC categories as novel classes and the remaining 60 categories are base classes. We sample the few-shot images following \cite{wang2020few,Sun_2021_CVPR}, and report the detection accuracy AP under shots 1, 2, 3, 5, 10 and 30 following \cite{Qiao_2021_ICCV,Han_2021_ICCV,wang2020few}. 
We report both single run results using the exact same few-shot images as \cite{wang2020few,kang2019few} and the average results of multiple runs.
We use the MSCOCO dataset under 2/10/30-shot for ablation study in Section \ref{ablation}.

\subsection{Implementation Details}
\label{imple_detail}

We implement our model based on the improved Pyramid Vision Transformer PVTv2 \cite{wang2021pvtv2}. We follow most of the model designs and hyperparameters in PVTv2.

The reason is that, \textbf{first}, PVTv2 is a pure transformer backbone, and has been shown strong performance on image classification, object detection, and etc. \textbf{Second}, the spatial-reduction attention (SRA) is initially proposed to reduce the computation overhead in PVT \cite{Wang_2021_ICCV} and PVTv2 \cite{wang2021pvtv2}. We find that it is also an effective way to summarize the key information in the high-resolution features. Inspired by this, we propose the asymmetric-batched cross-attention which aggregates the key information from the two branches for attention, using the sub-sampled features.

For experiments, we use the PVTv2 model variants PVTv2-B0, PVTv2-B1, PVTv2-B2 and PVTv2-B2-Li for implementation. We do not use PVTv2-B3 or larger models due to the GPU memory limit. Our model is initialized from the ImageNet pretrained model provided by \cite{wang2021pvtv2}. We use PVTv2-B2-Li as the default model because it can largely reduce the training/testing time using the pooling based spatial reduction attention, and maintains high detection accuracy.

The detailed training hyperparameters (\eg, epochs, learning rate) are included in the supplementary file.

\subsection{Ablation Study}
\label{ablation}

We perform ablation study on the model architecture and training strategy in Table \ref{tab:main_ablation}, \ref{tab:interaction}, and \ref{tab:training_method}.

\textbf{Single-branch baseline model versus Two-branch baseline model.} First, we compare the single-branch baseline model \cite{wang2020few} and the two-branch baseline model \cite{fan2020few} in Table \ref{tab:main_ablation} (a-d). We compare the performance of the two models using two feature backbones, ResNet-101 and PVTv2-B2-Li. Using the stronger transformer backbone, we achieve much higher FSOD accuracy. The two-branch based model outperforms the single-branch one using any of the two backbones, especially for extremely few-shot settings, \eg, 2/10-shot. The reason is that the single-branch based model is prone to overfitting to the few-shot training data, while the two-branch based model has stronger generalization ability by learning to compare the query regions with few-shot classes.

\begin{figure}[t]
\begin{center}
\includegraphics[scale=0.24]{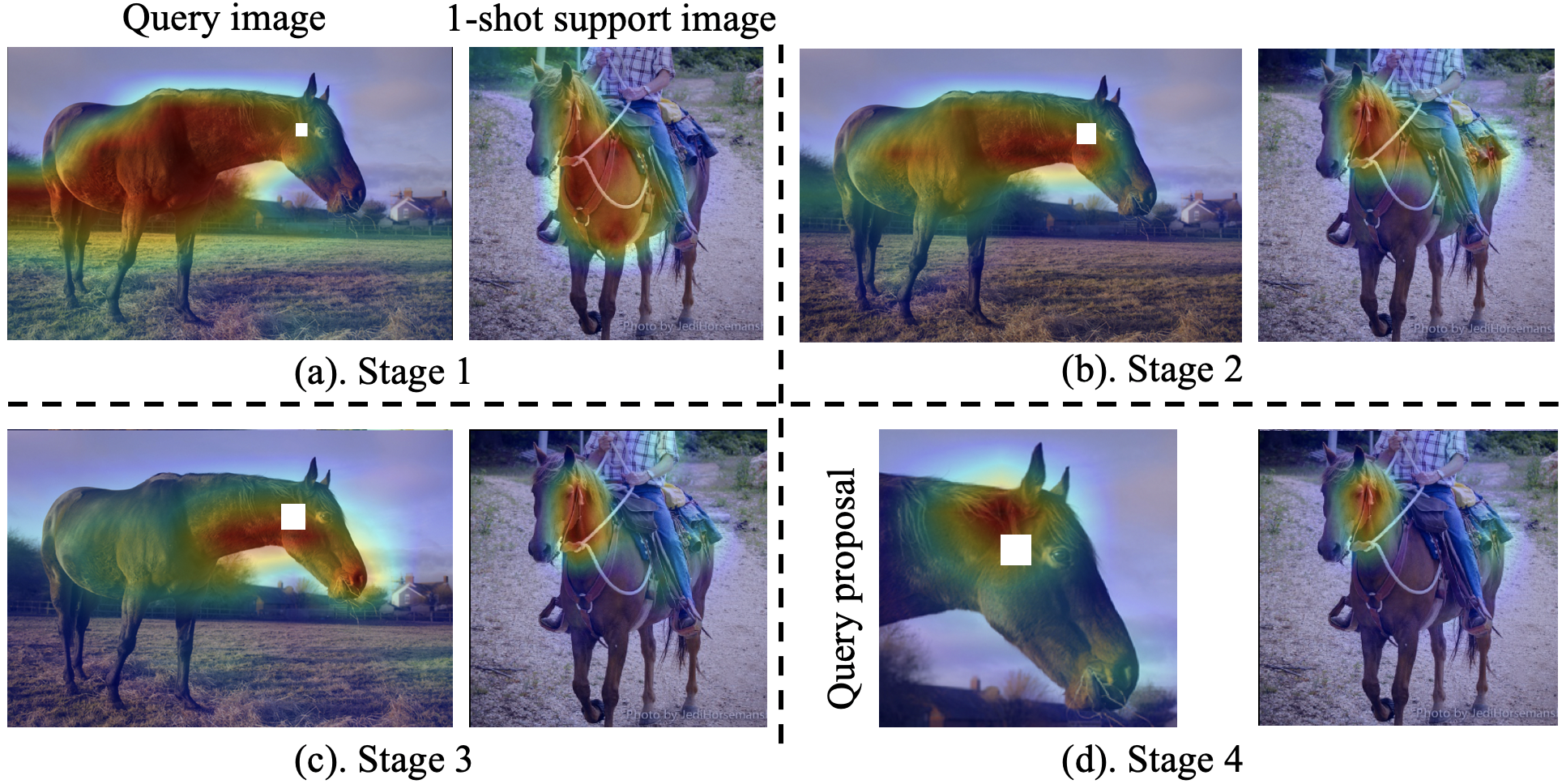}
\end{center}
\vspace{-2mm}
\caption{Visualization of the multi-level cross-attention in our model (\textcolor{red}{RED} means larger value). Using the white-box area (near the eye of the horse in the query) as Q, we show the corresponding cross-attention masks in both the query image and 1-shot support image. We visualize the last cross-transformer layer in all the four stages. The white boxes with different sizes in each stage are determined by the actual patch sizes in the input.}
\vspace{-2mm}
\label{figure_4}
\end{figure}

\textbf{How do each of the cross-transformer blocks help for FSOD?} We study the functions of the four cross-transformer stages in Table \ref{tab:main_ablation} (e-j). (1) We conduct the experiments of using only one cross-transformer stage and leave the other three stages for separate processing in Table \ref{tab:main_ablation} (e-h). The results show the effectiveness of all the four cross-transformer stages due to the mutual alignment of the two branches and feature fusion. In all four stages, Stage 4 in the detection head improves the most. This is because the objective of FSOD is to compare the proposal features with the support features, and Stage 4 unifies the RoI feature extraction of the two branches before the final comparison. (2) Using the first three stages results in our cross-transformer feature backbone (Table \ref{tab:main_ablation} (i)), which further improves the performance, compared with using any of these stages alone. Finally, our fully cross-transformer (FCT) (Table \ref{tab:main_ablation} (j)) achieves the best results with the cross-transformer feature backbone and detection head. (3) The visualization of the cross-attention masks in the four stages is shown in Figure \ref{figure_4}. From Figure~\ref{figure_4}, we have the following observations: i) In the early stages (\eg, stage 1), the attention masks spread out over the regions with similar color and texture, which align the low-level feature spaces of the two branches. ii) In the later stages, the attention masks focus more on semantic related local regions, which align the two high-level feature spaces.

\begin{table*}[ht]
    \centering
    \footnotesize
    % \vspace{-4mm}
    \caption{Ablation study on each component in our model using various backbones, tested on the MSCOCO dataset. $^{\dag}$ We replace the original block in the backbone with our cross-transformer block if marked. $^{\ddag}$ The baseline model has no cross-branch interaction in the feature backbone and RoI feature extractor. \vspace{-2mm}}
    \adjustbox{width=\linewidth}{
    \begin{tabular}{c|c|cccc|ccc|ccc|ccc}
    \toprule
    & \multirow{2}{*}{Backbone} & \multicolumn{4}{c|}{Our Cross-Transformer$^{\dag}$}
    & \multicolumn{3}{c|}{2-shot} & \multicolumn{3}{c|}{10-shot} & \multicolumn{3}{c}{30-shot} \\
    & & Stage 1 & {Stage 2} & {Stage 3} & {Stage 4}
    & AP & AP50 & AP75 & AP & AP50 & AP75 & AP & AP50 & AP75 \\ \midrule
    (a) & ResNet101 & \multicolumn{4}{c|}{Single branch baseline model \cite{wang2020few}$^{\ddag}$} & 4.6 & 8.3 & 4.8   & 10.0 & 19.1 & 9.3   & 13.7 & 24.9 & 13.4 \\
    (b) & ResNet101 & \multicolumn{4}{c|}{Two branch baseline model \cite{fan2020few}$^{\ddag}$} & 5.6 & 14.0 & 3.9   & 9.6 & 20.7 & 7.7   & 13.5 & 28.5 & 11.7 \\ \midrule
    (c) & PVTv2-B2-Li & \multicolumn{4}{c|}{Single branch baseline model \cite{wang2020few}$^{\ddag}$} & 5.3 & 9.5 & 5.2   & 14.5 & 26.5 & 13.9   & 19.7 & 33.6 & 19.9 \\
    (d) & PVTv2-B2-Li & \multicolumn{4}{c|}{Two branch baseline model \cite{fan2020few}$^{\ddag}$} & 7.0 & 12.8 & 6.7   & 15.3 & 27.3 & 15.3   & 19.5 & 32.7 & 19.8 \\ \midrule
    (e) & PVTv2-B2-Li & \checkmark & & & & 7.1 & 13.0 & 6.8   & 15.7 & 28.3 & 15.4   & 20.2 & 33.6 & 20.5 \\
    (f) & PVTv2-B2-Li & & \checkmark & & & 7.3 & 13.1 & 7.0   & 16.2 & 28.5 & 16.0   & 20.4 & 33.9 & 20.8 \\
    (g) & PVTv2-B2-Li & & & \checkmark & & 7.4 & 13.3 & 7.3   & 16.1 & 28.5 & 15.8   & 20.5 & 33.8 & 20.9 \\
    (h) & PVTv2-B2-Li & & & & \checkmark & 7.7 & 13.5 & 7.7   & 16.4 & 28.9 & 16.3   & 20.7 & 34.1 & 21.5 \\ 
    (i) & PVTv2-B2-Li & \checkmark & \checkmark & \checkmark & & 7.6 & 13.7 & 7.6  & 16.5 & 29.6 & 16.2   &  20.8 & 34.9 & 21.2 \\ 
    (j) & PVTv2-B2-Li & \checkmark & \checkmark & \checkmark & \checkmark & \textbf{7.9} & \textbf{14.2} & \textbf{7.9}   & \textbf{17.1} & \textbf{30.2} & \textbf{17.0}   & \textbf{21.4} & 35.5 & \textbf{22.1} \\ \midrule
    (k) & PVTv2-B0 & \checkmark & \checkmark & \checkmark & \checkmark & 4.6 & 8.1 & 4.2   & 10.2 & 20.1 & 8.7   & 13.7 & 27.5 & 11.8 \\ 
    (l) & PVTv2-B1 & \checkmark & \checkmark & \checkmark & \checkmark & 5.3 & 9.5 & 5.0   & 12.1 & 23.9 & 10.2   & 17.3 & 33.4 & 15.6 \\ 
    (m) & PVTv2-B2 & \checkmark & \checkmark & \checkmark & \checkmark & 7.3 & 13.7 & 7.2   & 16.3 & 29.6 & 16.4   & 20.6 & \textbf{37.2} & 20.8 \\
    \bottomrule
    \end{tabular}}
\label{tab:main_ablation}
\vspace{-2mm}
\end{table*}

\begin{table}[t]
    \centering
    \footnotesize
    \caption{Ablation study on the aggregation of the key-value pairs from the two branches. \vspace{-2mm}}
    \adjustbox{width=\linewidth}{
    \begin{tabular}{l|ccc|ccc}
    \toprule
    \multirow{2}{*}{Method} & \multicolumn{3}{c|}{2-shot} & \multicolumn{3}{c}{10-shot}\\
    & AP & AP50 & AP75 & AP & AP50 & AP75 \\ \midrule
    Addition & 6.5 & 11.9 & 6.2  & 15.0 & 26.2 & 14.8 \\
    Multiplication & 6.7 & 12.0 & 6.7  & 15.1 & 26.9 & 15.0 \\ \midrule
    W/o branch embed & 7.7 & 14.0 & 7.8  & 17.0 & 29.8 & 17.0 \\ 
    W/ branch embed & \textbf{7.9} & \textbf{14.2} & \textbf{7.9}   & \textbf{17.1} & \textbf{30.2} & \textbf{17.0}  \\
    \bottomrule
    \end{tabular}}
\label{tab:interaction}
\end{table}

\begin{table}[t]
    \centering
    \footnotesize
    \caption{Ablation study on model training framework. \vspace{-2mm}}
    \adjustbox{width=\linewidth}{
    \begin{tabular}{c|ccc|ccc}
    \toprule
    {Single-branch} & \multicolumn{3}{c|}{2-shot} & \multicolumn{3}{c}{10-shot}\\
    pretraining & AP & AP50 & AP75 & AP & AP50 & AP75 \\ \midrule
    & 5.3 & 10.3 & 5.0  & 14.1 & 25.5 & 13.3 \\ 
    \checkmark & \textbf{7.9} & \textbf{14.2} & \textbf{7.9}   & \textbf{17.1} & \textbf{30.2} & \textbf{17.0}  \\
    \bottomrule
    \end{tabular}}
\label{tab:training_method}
\vspace{-2mm}
\end{table}

\textbf{The comparison of model performance using different backbones.} We conduct the experiments using different PVTv2 variants as the backbone in Table \ref{tab:main_ablation} (j-m). The PVTv2-B2 based model outperforms the models based on PVTv2-B0 and PVTv2-B1 due to larger model capacity. The PVTv2-B2-Li based model has very similar performance compared with PVTv2-B2, and is faster for training/testing speed. Therefore, we use PVTv2-B2-Li by default.

\textbf{The ablation study on the information aggregation across branches.} To perform cross-attention with the two branches, we need to aggregate the key information from both of them. Specifically, we use the concatenation operation with branch embedding to aggregate the K-V pairs from the two branches in our work, without losing the original information. (1) We conduct the experiments using the element-wise addition and multiplication for aggregating the K-V pairs of the two branches. The results are much worse compared with using the concatenation, as shown in Table \ref{tab:interaction}, due to the potential information loss. (2) The branch embedding can identify which branch the feature comes from, and slightly improve the performance in Table \ref{tab:interaction}. 

\textbf{The importance of the three-step training framework.} We have three steps for model training. The first and second step are both pre-training, performed over the data-abundant base classes. We conduct the experiments of using the first-step pre-training or not in Table \ref{tab:training_method}. Using the single-branch pre-training leads to a large improvement. This is because the single-branch method with a multi-class classifier is good at learning a stronger feature backbone over large-scale base-class training data, while our two-branch based method is better for the few-shot scenario by learning how to compare. Therefore, we combine the strengths of the two methods in the first two steps of the training. The pre-trained model in the first step can provide a good initialization, which can help ease the training in the second step.

\begin{table*}[ht]
\centering
\footnotesize
\setlength{\tabcolsep}{0.4em}
\caption{{Few-shot object detection results (AP50) on the PASCAL VOC dataset. We report both single run results and the average results of multiple runs. S: Single-branch based methods. T: Two-branch based methods. \vspace{-2mm}}} 
\adjustbox{width=\linewidth}{
\begin{tabular}{c|c|c|c|ccccc|ccccc|ccccc}
\toprule
\multirow{2}{*}{Type} & \multirow{2}{*}{Method} & \multirow{2}{*}{Venue} & \multirow{2}{*}{Backbone} & \multicolumn{5}{c|}{Novel Set 1} & \multicolumn{5}{c|}{Novel Set 2} & \multicolumn{5}{c}{Novel Set 3} \\ 
& &  & & 1     & 2     & 3    & 5    & 10   & 1     & 2     & 3    & 5    & 10   & 1     & 2     & 3    & 5    & 10   \\ \midrule

\multicolumn{19}{c}{\textbf{Single run results, using the exact same few-shot samples as \cite{wang2020few}}} \\ \midrule
\multirow{6}{*}{S} & MetaDet~\cite{wang2019meta} & ICCV 2019 & VGG16 & 18.9 & 20.6 & 30.2 & 36.8 & 49.6 & 21.8 & 23.1 & 27.8 & 31.7 & 43.0 & 20.6 & 23.9 & 29.4 & 43.9 & 44.1 \\ 
& TFA w/ cos \cite{wang2020few} & ICML 2020 & ResNet-101 & 39.8 & 36.1 & 44.7 & 55.7 & 56.0 & 23.5 & 26.9 & 34.1 & 35.1 & 39.1 & 30.8 & 34.8 & 42.8 & 49.5 & 49.8 \\ 
& MPSR \cite{wu2020multi} & ECCV 2020 & ResNet-101 & 41.7 & 42.5 & 51.4 & 55.2 & 61.8 & 24.4 & 29.3 & 39.2 & 39.9 & 47.8 & 35.6 & 41.8 & 42.3 & 48.0 & 49.7 \\ 
& SRR-FSD \cite{Zhu_2021_CVPR} & CVPR 2021 & ResNet-101 & 47.8 & 50.5 & 51.3 & 55.2 & 56.8    & 32.5 & 35.3 & 39.1 & 40.8 & 43.8    & 40.1 & 41.5 & 44.3 & 46.9 & 46.4 \\
& CoRPNs + Halluc \cite{Zhang_2021_CVPR} & CVPR 2021 & ResNet-101 & 47.0 & 44.9 & 46.5 & 54.7 & 54.7   & 26.3 & 31.8 & 37.4 & 37.4 & 41.2   & 40.4 & 42.1 & 43.3 & 51.4 & 49.6 \\
& FSCE \cite{Sun_2021_CVPR} & CVPR 2021 & ResNet-101 & 44.2 & 43.8 & 51.4 & 61.9 & 63.4    & 27.3 & 29.5 & 43.5 & 44.2 & 50.2    & 37.2 & 41.9 & 47.5 & 54.6 & 58.5 \\ \hline
\multirow{5}{*}{T} & FSRW~\cite{kang2019few}  & ICCV 2019 & YOLOv2 & 14.8  & 15.5  & 26.7 & 33.9 & 47.2 & 15.7  & 15.3  & 22.7 & 30.1 & 40.5 & 21.3  & 25.6  & 28.4 & 42.8 & 45.9 \\ 
& Meta R-CNN~\cite{yan2019meta} & ICCV 2019 & ResNet-101 & 19.9 & 25.5 & 35.0 & 45.7 & 51.5 & 10.4 & 19.4 & 29.6 & 34.8 & 45.4 & 14.3 & 18.2 & 27.5 & 41.2 & 48.1 \\ 
& Fan et al. \cite{fan2020few} & CVPR 2020 & ResNet-101 & 37.8 & 43.6 & 51.6 & 56.5 & 58.6    & 22.5 & 30.6 & 40.7 & 43.1 & 47.6    & 31.0 & 37.9 & 43.7 & 51.3 & 49.8 \\
& QA-FewDet \cite{Han_2021_ICCV} & ICCV 2021 & ResNet-101 & 42.4 & 51.9 & 55.7 & 62.6 & 63.4 & 25.9 & \textbf{37.8} & \textbf{46.6} & 48.9 & 51.1  & 35.2 & 42.9 & 47.8 & 54.8 & 53.5 \\ 
& Meta Faster R-CNN \cite{han2021meta} & AAAI 2022 & ResNet-101 & 43.0 & {54.5} & \textbf{60.6} & \textbf{66.1} & {65.4}   & 27.7 & {35.5} & {46.1} & {47.8} & \textbf{51.4}   & \textbf{40.6} & {46.4} & \textbf{53.4} & \textbf{59.9} & {58.6}\\
& FCT (Ours) & This work & PVTv2-B2-Li & \textbf{49.9} & \textbf{57.1} & {57.9} & {63.2} & \textbf{67.1}    & \textbf{27.6} & 34.5 & 43.7 & \textbf{49.2} & {51.2}     & {39.5} & \textbf{54.7} & {52.3} & {57.0} & \textbf{58.7} \\ \midrule
\multicolumn{19}{c}{\textbf{Average results of multiple runs, following \cite{wang2020few}}} \\ \midrule
\multirow{3}{*}{S} & TFA w/ cos \cite{wang2020few} & ICML 2020 & ResNet-101 & 25.3 & 36.4 & 42.1 & 47.9 & 52.8  & 18.3 & 27.5 & 30.9 & 34.1 & 39.5  & 17.9 & 27.2 & 34.3 & 40.8 & 45.6 \\
& FSCE \cite{Sun_2021_CVPR} & CVPR 2021 & ResNet-101 & 32.9 & 44.0 & 46.8 & 52.9 & 59.7 & 23.7 & 30.6 & 38.4 & 43.0 & 48.5 & 22.6 & 33.4 & 39.5 & 47.3 & 54.0 \\
& DeFRCN \cite{Qiao_2021_ICCV} & ICCV 2021 & ResNet-101 & {40.2} & {53.6} & {58.2} & {63.6} & {66.5}    & {29.5} & {39.7} & {43.4} & {48.1} & {52.8}   & {35.0} & 38.3 & {52.9} & {57.7} & {60.8} \\ \hline
\multirow{3}{*}{T} & Xiao et al. \cite{xiao2020few} & ECCV 2020 & ResNet-101 & 24.2 & 35.3 &  42.2 &  49.1 &  57.4 & 21.6 & 24.6 &  31.9 &  37.0 &  45.7 & 21.2 &  30.0 &  37.2 &  43.8 &  49.6 \\
& DCNet \cite{Hu_2021_CVPR} & CVPR 2021 & ResNet-101 & 33.9 & 37.4 & 43.7 & 51.1 & 59.6 & 23.2 & 24.8 & 30.6 & 36.7 & 46.6 & 32.3 & 34.9 & 39.7 & 42.6 & 50.7 \\
& FCT (Ours) & This work & PVTv2-B2-Li & \textbf{38.5} & \textbf{49.6} & \textbf{53.5} & \textbf{59.8} & \textbf{64.3}  & \textbf{25.9} & \textbf{34.2} & \textbf{40.1} & \textbf{44.9} & \textbf{47.4}    & \textbf{34.7} & \textbf{43.9} & \textbf{49.3} & \textbf{53.1} & \textbf{56.3} \\ 
\bottomrule
\end{tabular}}
\label{tab:main_voc}
\vspace{-2mm}
\end{table*}

\begin{table}[ht]
\centering
\footnotesize
\setlength{\tabcolsep}{2.0mm}
\caption{{Few-shot object detection results (AP) on the MSCOCO dataset. S: Single-branch based methods. T: Two-branch based methods. \vspace{-2mm}}}
\adjustbox{width=\linewidth}{
\begin{tabular}{c|c|cccccc}
\toprule
\multirow{2}{*}{Type} & \multirow{2}{*}{Method} & \multicolumn{6}{c}{Shot} \\
& & 1 & 2 & 3 & 5 & 10 & 30 \\ \midrule
\multicolumn{8}{c}{\textbf{Single run results, using the exact same few-shot samples as \cite{wang2020few}}} \\ \midrule
\multirow{6}{*}{S} & MetaDet\small{~\cite{wang2019meta}} & {--} & {--} & {--} & {--} & 7.1 & 11.3 \\
& TFA w/ cos \cite{wang2020few} & 3.4  & 4.6  & 6.6 & 8.3 & 10.0 & 13.7 \\
& MPSR \cite{wu2020multi}       & {2.3}  & {3.5} & {5.2} &{6.7} & {9.8} & {14.1} \\
& SRR-FSD \cite{Zhu_2021_CVPR}  & - & - & - & - & 11.3 & 14.7 \\
& TFA + Halluc \cite{Zhang_2021_CVPR} & 4.4 & 5.6 & 7.2 & - & - & - \\
& FSCE \cite{Sun_2021_CVPR} & - & - & - & - & 11.9 & 16.4 \\ \hline
\multirow{5}{*}{T} & FSRW\small{~\cite{kang2019few}}  & {--} & {--} & {--} & {--} & 5.6 & 9.1 \\
& Meta R-CNN \cite{yan2019meta} & {--} & {--} & {--} & {--} & {8.7} & {12.4} \\
& Fan et al. \cite{fan2020few} & 4.2 & 5.6 & 6.6 & 8.0 & 9.6 & 13.5 \\
& QA-FewDet \cite{Han_2021_ICCV} & 4.9  & 7.6 & 8.4 & 9.7 & 11.6 & 16.5 \\
& Meta Faster R-CNN \cite{han2021meta} & {5.1} & {7.6} & {9.8} & {10.8} & {12.7} & {16.6} \\
& FCT (Ours) & \textbf{5.6} & \textbf{7.9} & \textbf{11.1} & \textbf{14.0} & \textbf{17.1} & \textbf{21.4} \\\midrule
\multicolumn{8}{c}{\textbf{Average results of multiple runs, following \cite{wang2020few}}} \\ \midrule
\multirow{3}{*}{S} & TFA w/ cos \cite{wang2020few}   & 1.9 & 3.9 & 5.1 & 7.0 & 9.1  & 12.1 \\
& FSCE \cite{Sun_2021_CVPR}       & - & - & - & - & 11.1 & 15.3 \\
& DeFRCN \cite{Qiao_2021_ICCV}    & 4.8 &  {8.5} &  {10.7} &  {13.6} &  {16.8} &  {21.2} \\ \hline
\multirow{3}{*}{T} & Xiao et al. \cite{xiao2020few}  & 4.5 & 6.6 & 7.2 & 10.7 & 12.5 & 14.7 \\
& DCNet \cite{Hu_2021_CVPR}       & - & - & - & - & 12.8 & 18.6 \\
& FCT (Ours) & \textbf{5.1} & \textbf{7.2} & \textbf{9.8} & \textbf{12.0} & \textbf{15.3} & \textbf{20.2} \\
\bottomrule
\end{tabular}}
\label{tab:main_coco}
\vspace{-2mm}
\end{table}

\subsection{Comparison with the State-of-the-arts (SOTAs)}

We compare our proposed FCT with the recent state-of-the-arts on the PASCAL VOC and MSCOCO FSOD benchmarks in Table \ref{tab:main_voc} and \ref{tab:main_coco}. We report both the single run and multiple runs results following \cite{wang2020few,Sun_2021_CVPR} on the two benchmarks. Compared with the existing two-branch based methods, we achieve the SOTAs across most of the shots under the two evaluation settings in the two benchmarks. 

Compared with the single-branch based methods, we achieve the second best results under the multiple runs setting.
DeFRCN \cite{Qiao_2021_ICCV}, reports the best results with multiple runs, which is a highly-optimized single-branch based method. It proposes a Gradient Decoupled Layer to adjust the degree of decoupling of the backbone, RPN, and R-CNN through gradient, and also a post-processing Prototypical Calibration Block. Different from that, we propose a novel two-branch based FSOD model, and achieves the best results on the most challenging MSCOCO 1-shot setting with multiple runs. This is because we do not learn the multi-class classifier over novel classes, and instead learn the class-agnostic comparison network between the query and support, which is shared among all classes. Thus, our method can mitigate the data scarcity problem under 1-shot setting and improve the model generalization ability.

\section{Conclusion}

We propose a novel fully cross-transformer based few-shot object detection model (FCT) in this work, by incorporating cross-transformer into both the feature backbone and detection head. The asymmetric-batched cross-attention is proposed to aggregate the K-V pairs from the query and support branch with different batch sizes. We show both quantitative results on the two widely used FSOD benchmarks and qualitative visualization of the multi-level cross-attention learned in our model. All these evidence demonstrates the effectiveness of the proposed multi-level interactions between the query and support branch. We hope our work can inspire future work on the two-branch based FSOD methods.

\section*{Acknowledgements}
{
\noindent This material is based on research sponsored by Air Force Research Laboratory (AFRL) under agreement number FA8750-19-1-1000. 
The U.S. Government is authorized to reproduce and distribute reprints for Government purposes notwithstanding any copyright notation therein. 
The views and conclusions contained herein are those of the authors and should not be interpreted as necessarily representing the official policies or endorsements, either expressed or implied, of Air Force Laboratory, DARPA or the U.S. Government.
}

%%%%%%%%% REFERENCES
{\small
\bibliographystyle{ieee_fullname}
\bibliography{egbib}
}

\clearpage
\newpage
\appendix

\section*{Appendix}
The supplementary materials are organized as follows. First, we describe the implementation details of our training framework in Section \ref{impls}. Then, we show more visualization examples of our multi-level cross-attention in Section \ref{visual}.

\section{Implementation Details for Model Training}
\label{impls}

We have three steps for model training. In the first two steps, the model is trained over the data-abundant base-classes dataset. In the last step, the model is fine-tuned over novel classes and then used for evaluation.

\textbf{(1) Pre-training the single-branch based model over base classes.} In the first step, we train the single-branch based few-shot object detection model using large-scale base-class dataset. For model architecture, we use the vallina Faster R-CNN model with the vision transformer backbone (PVTv2 \cite{Wang_2021_ICCV,wang2021pvtv2} in this work). The model is initialized from the ImageNet pretrained model provided by \cite{wang2021pvtv2}. We follow the original Faster R-CNN paper for model training. The loss function is defined as,
\begin{equation}
\mathcal{L}_1 = \mathcal{L}_{rpn} + \mathcal{L}_{rcnn}
\end{equation}
where $\mathcal{L}_{rpn}$ and $\mathcal{L}_{rcnn}$ both consist of a classification loss and a bbox regression loss as follows,
\begin{equation}
\mathcal{L}_{rpn} = \mathcal{L}^{B}_{cls} + \mathcal{L}_{loc},\;\;\; \mathcal{L}_{rcnn} = \mathcal{L}^{M}_{cls} + \mathcal{L}_{loc}
\end{equation}
where $\mathcal{L}^{B}_{cls}$ denotes the binary cross-entropy loss over a “foreground" class (the union of all base classes) and a “background" class, and $\mathcal{L}^{M}_{cls}$ denotes the multi-class cross-entropy loss over all base classes plus a “background” class. The $\mathcal{L}_{loc}$ denotes the bbox regression loss using smooth $L_1$ loss defined in \cite{Fast_R-CNN}.

For model training on the MSCOCO dataset, we use the AdamW optimizer with an initial learning rate of 0.0002, weight decay of 0.0001, and a batch size of 8. The learning rate is divided by 10 after 85,000 and 100,000 iterations. The total number of training iterations is 110,000.

Similarly, we use smaller training iterations for model training on the PASCAL VOC dataset. The initial learning rate is 0.0002, divided by 10 after 12,000 and 16,000 iterations. The total number of training iterations is 18,000.

\textbf{(2) Training the two-branch based model over base classes.} In this step, we train the proposed two-branch based model with fully cross-transformer (FCT). The model is initialized by the pretrained model in the first step. Our FCT model can reuse most of the parameters of the single-branch based model in the first step, and only need to learn the branch embedding and pairwise matching network \cite{fan2020few}. We also show in Table \ref{tab:training_method} of the main paper the importance of the pre-trained single-branch based model. The loss function is defined as,

\begin{equation}
\label{joint_loss}
\mathcal{L}_2 = \mathcal{L}_{att\_rpn} + \mathcal{L}_{matching}
\end{equation}
where $\mathcal{L}_{att\_rpn}$ and $\mathcal{L}_{matching}$ both consist of a binary cross-entropy loss and a bbox regression loss,
\begin{equation}
\mathcal{L}_{att\_rpn} = \mathcal{L}^{B}_{cls} + \mathcal{L}_{loc},\;\;\; \mathcal{L}_{matching} = \mathcal{L}^{B}_{cls} + \mathcal{L}_{loc}
\end{equation}
where the Attention-RPN and the pairwise matching network both use the binary cross-entropy loss $\mathcal{L}^{B}_{cls}$ and bbox regression loss $\mathcal{L}_{loc}$ for training, following \cite{fan2020few}. 

For model training on the MSCOCO dataset, we use the AdamW optimizer with an initial learning rate of 0.0001, weight decay of 0.0002, and a batch size of 4. The learning rate is divided by 10 after 15,000 and 20,000 iterations. The total number of training iterations is 20,000. We use much smaller iterations than in the first step thanks to the good initialization.

For model training on the PASCAL VOC dataset, we use the same hyper-parameters as on the MSCOCO dataset except using fewer training iterations. The initial learning rate is 0.0002, divided by 10 after 7,500 and 10,000 iterations. The total number of training iterations is 10,000.

\textbf{(3) Fine-tuning the two-branch based model over novel classes.} In this step, the model is fine-tuned using a sub-sampled $K$-shot dataset with both base classes and novel classes. We use the same loss function in the second step for model training. After the few-shot fine-tuning, the learned model is used for evaluation.

\begin{figure*}
\begin{center}
\includegraphics[scale=0.68]{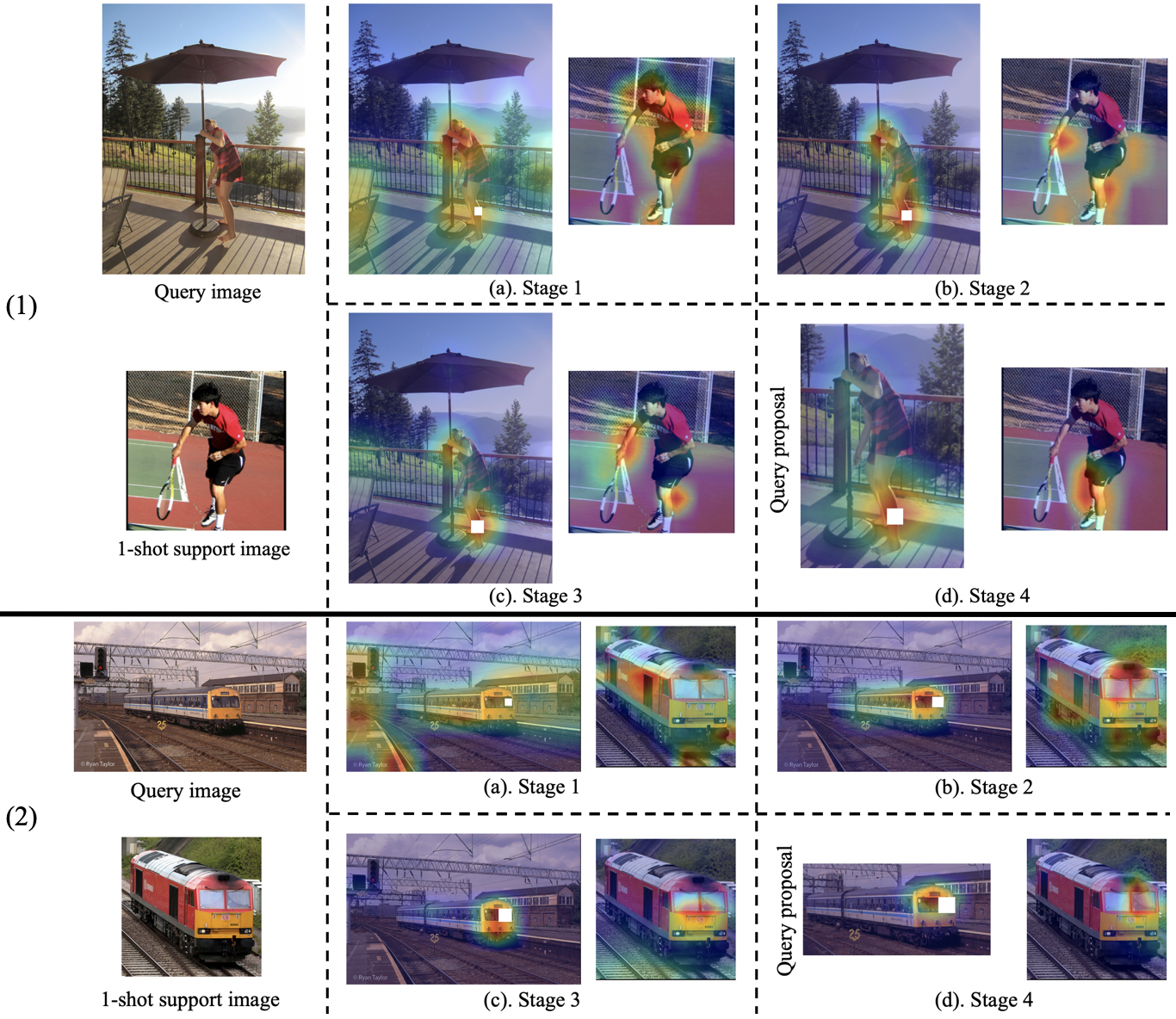}
\end{center}
\caption{Visualization of the multi-level cross-attention, similar as the Figure \ref{figure_4} in the main paper.}
\label{figure_5}
\end{figure*}

For model training on the MSCOCO and PASCAL VOC dataset, we use the AdamW optimizer with an initial learning rate of 0.0002, weight decay of 0.0001, and a batch size of 4. For 30-shot fine-tuning, the learning rate is divided by 10 after 3,000 iterations, and the total number of training iterations is 5,000. For 10-shot fine-tuning or fewer, the learning rate is divided by 10 after 2,000 iterations, and the total number of training iterations is 3,000.

\section{Visualization of Multi-level Cross-attention}
\label{visual}

\begin{figure*}
\begin{center}
\includegraphics[scale=0.68]{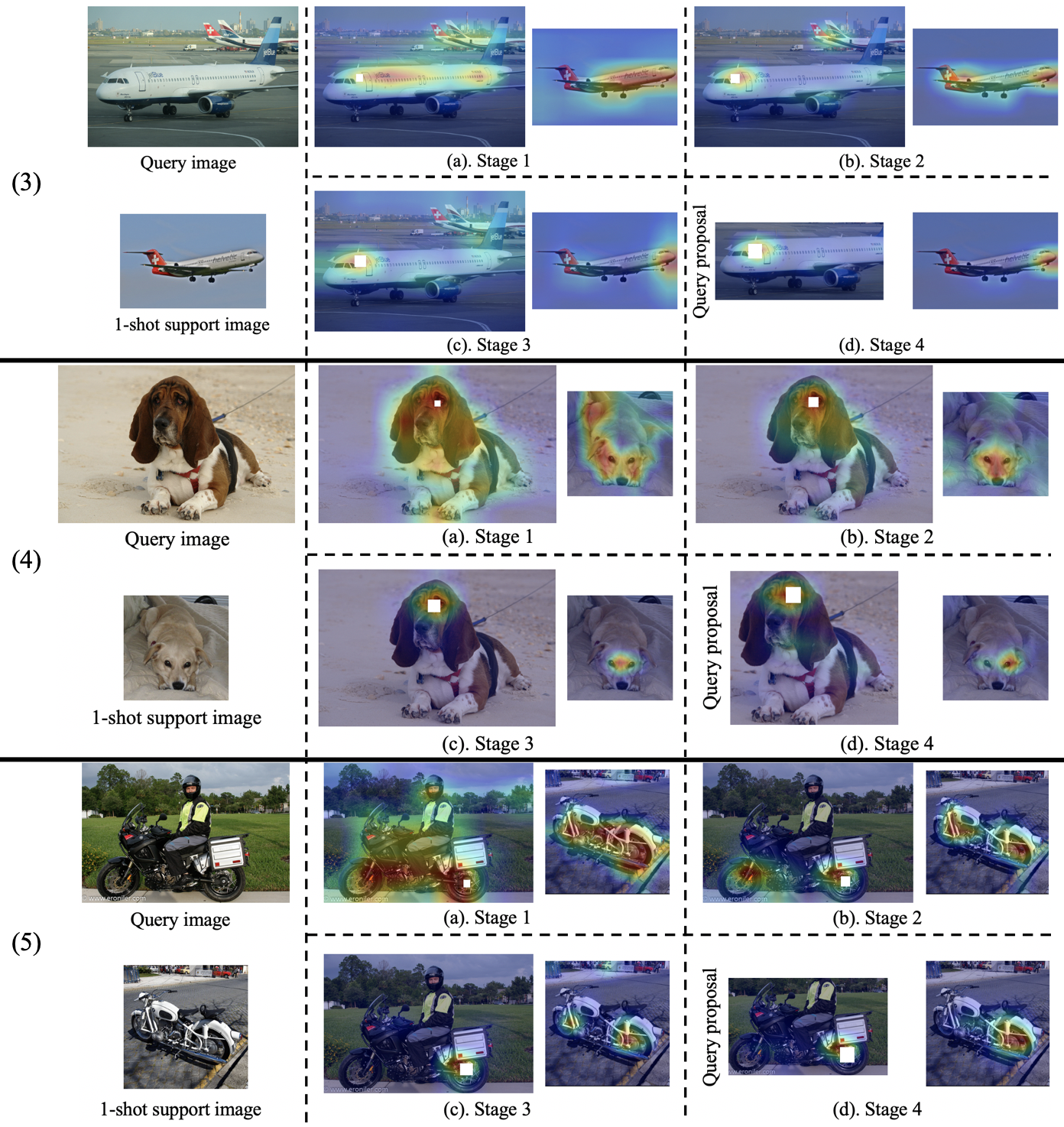}
\end{center}
\caption{Visualization of the multi-level cross-attention, similar as the Figure \ref{figure_4} in the main paper.}
\label{figure_6}
\end{figure*}

We show more visualization examples of the proposed multi-level cross-attention in Figure \ref{figure_5} and \ref{figure_6}.

\end{document}